\documentclass[11pt]{article}

\usepackage[final]{acl}

\usepackage{times}
\usepackage{latexsym}
\usepackage{booktabs}
\usepackage{amsmath}
\usepackage{float}

\usepackage[T1]{fontenc}

\usepackage[utf8]{inputenc}

\usepackage{microtype}

\usepackage{inconsolata}

\usepackage{graphicx}

\title{CSE-UOI at SemEval-2026 Task 6: A Two-Stage Heterogeneous Ensemble with Deliberative Complexity Gating for Political Evasion Detection}

\author{
  Christos Tzouvaras\textsuperscript{1}, 
  Konstantinos Skianis\textsuperscript{1}, 
  Athanasios Voulodimos\textsuperscript{2} \\
  \textsuperscript{1}University of Ioannina \\
  \textsuperscript{2}National Technical University of Athens \\
  \texttt{contact@tzouvarasc.com, kskianis@cse.uoi.gr} \\
  \texttt{thanosv@mail.ntua.gr}
}

\begin{document}
\maketitle
\begin{abstract}
  This paper describes our system for SemEval-2026 Task 6, which classifies clarity of responses in political interviews into three categories: Clear Reply, Ambivalent, and Clear Non-Reply. We propose a heterogeneous dual large language model (LLM) ensemble via self-consistency (SC) and weighted voting, and a novel post-hoc correction mechanism, Deliberative Complexity Gating (DCG). This mechanism uses cross-model behavioral signals and exploits the finding that an LLM response-length proxy correlates strongly with sample ambiguity. To further examine mechanisms for improving ambiguity detection, we evaluated multi-agent debate as an alternative strategy for increasing deliberative capacity. Unlike DCG, which adaptively gates reasoning using cross-model behavioral signals, debate increases agent count without increasing model diversity. Our solution achieved a Macro-F1 score of 0.85 on the evaluation set, securing 3rd place and tied with the second-best reported score.
\end{abstract}

\section{Introduction}

Political interviews frequently contain responses that avoid directly addressing posed questions \cite{bull2003microanalysis}, making the automatic identification of evasive or ambiguous answers an important problem for computational discourse analysis and political communication research. 
\citet{thomas2024isaidthatdataset} has framed this problem as response clarity classification for English, requiring models to distinguish between clear replies, ambivalent responses, and explicit non-replies under subtle pragmatic variation, leading to the current SemEval Task 6 \cite{thomas2026semeval2026task6clarity}.

In this work, we present a two-stage heterogeneous ensemble, combining cross-model self-consistency (SC) with a novel post-hoc mechanism, Deliberative Complexity Gating (DCG), which adaptively adjusts decision confidence using behavioral signals derived from model responses, including response-length-based ambiguity indicators. 
Unlike approaches that increase reasoning through additional agents or iterative debate, our method leverages model diversity and lightweight deliberation control to improve robustness in ambiguity detection.

Experimental results show that this design achieves strong performance for political interview clarity classification, highlighting the effectiveness of adaptive deliberation and heterogeneous ensembles, achieving Macro-F1 of 0.85 for Subtask 1, ranking 3rd out of 41 submissions and tied with the second-best reported score.
Our system's code is publicly available\footnote{\url{https://github.com/tzouvarasc/CSE-UOI-at-SemEval-2026-Task-6}}.

\begin{figure*}[!t]
  \centering
  \includegraphics[width=\textwidth]{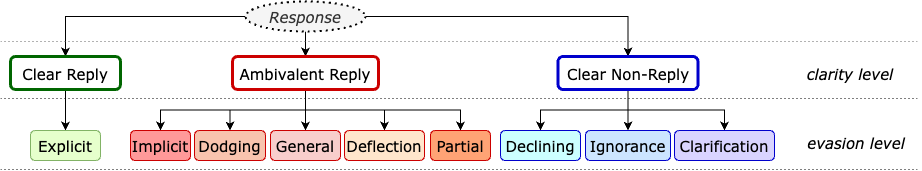}
  \caption{Task label taxonomy by \citet{thomas2024isaidthatdataset} (evasion labels mapped to 3 clarity classes).}
  \label{fig:taxonomy}
\end{figure*}

\section{Background}

\subsection{Task Setup}
SemEval-2026 Task 6 studies English political interview response clarity.
Each sample is a question--answer pair, and the output is one of three labels:
\textit{Clear Reply}, \textit{Ambivalent}, or \textit{Clear Non-Reply}.
We participate in Subtask 1 (clarity classification).
Our experiments use the official labeled split ($n=308$) for development and the blind evaluation split ($n=237$) for final evaluation.
Figure~\ref{fig:taxonomy} shows the two-level taxonomy used in our pipeline, mapping fine-grained evasion behaviors to the three clarity classes of Subtask~1.

\subsection{Related Work}

Recent work has examined behavioral signals in large language models (LLMs) as proxies for uncertainty and response reliability. Prior studies suggest that response length, in particular, correlates with ambiguity and model confidence during generation. \citet{janiak2025illusion} demonstrates that hallucination detection performance is strongly influenced by verbosity effects, showing that incorrect or uncertain responses often exhibit distinctive length patterns and higher variance. 
Similarly, \citet{zhao2025does} reports that factual accuracy tends to degrade as responses grow longer, attributing this phenomenon to knowledge exhaustion and confidence decay during extended generation.
Moreover, \citet{dong2025emergent} shows that models internally encode global response properties such as anticipated reasoning depth and output length, stating that generation length reflects latent complexity signals.

A substantial line of research studies adaptive inference methods that allocate additional computation only when needed. Early-exit language modeling \cite{schuster2022confident, zhang2025confidence} dynamically halt computation based on confidence thresholds, reducing compute on simpler continuations while preserving output quality.
Related early-exiting frameworks further refine token-level exit decisions and analyze robustness and performance tradeoffs under different gating criteria \cite{bae2023fast}.
Another line of work increases deliberative capacity through multi-agent debate and collaborative reasoning, where multiple language model instances iteratively critique and refine candidate answers to improve reliability and factual consistency \cite{du2024improving, liang2024encouraging, hu2025multiagent}.
These approaches collectively motivate DCG as a lightweight post-hoc strategy for adaptively modulating deliberative complexity using observable cross-model behavioral cues.
Our contribution differs from prior approaches by introducing a transductive, post-hoc gating layer that uses cross-model behavioral signals without extra training or additional API calls at Stage 2.

\begin{figure*}[t]
  \centering
  \includegraphics[width=\textwidth]{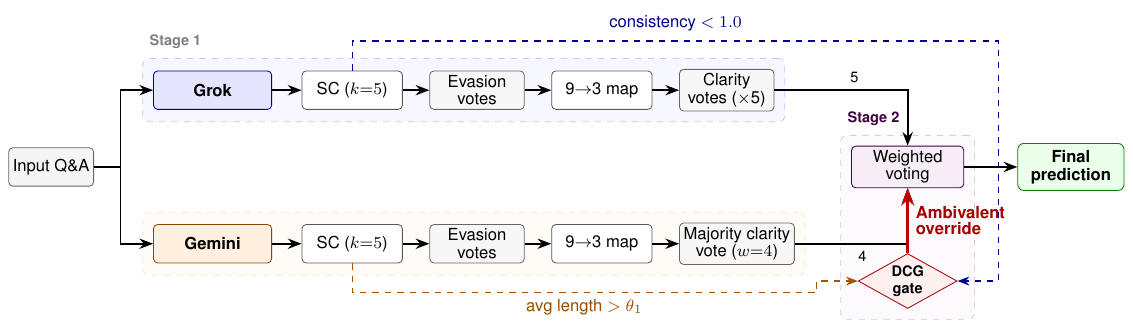}
  \caption{Architecture of our two-stage pipeline. Stage 1 runs Grok and Gemini with $k=5$ self-consistency (SC), maps evasion to clarity, and applies asymmetric weighted voting. Stage 2 (DCG) uses Gemini response length and Grok consistency to gate uncertain cases.}
  \label{fig:architecture}
\end{figure*}

\section{System Overview}

\subsection{Dual-Model Ensemble}

Our base system, which we refer to as Stage 1, is a heterogeneous dual-model ensemble. Instead of directly predicting the three clarity classes, we use an evasion-first strategy. The models are instructed to classify responses into the task's nine fine-grained evasion categories. The resulting evasion label is then deterministically mapped to \{Clear Reply, Ambivalent, Clear Non-Reply\}. This mapping acts as an error buffer. From development and evaluation set analysis combined, 58.5\% of evasion-level errors remain within the same clarity bucket and therefore preserve the final clarity prediction. Evasion-level performance broken down per class is reported in (Appendix~\ref{app:c3}).

We run \texttt{grok-4-1-fast-reasoning} and \texttt{gemini-3-flash-preview} in parallel with self-consistency ($k=5$) for each model \cite{wang2023selfconsistency}. Both models are prompted with the same structured evasion taxonomy Chain-of-Thought (CoT) prompt \cite{wei2022chain}.

To aggregate predictions, we use an asymmetric weighted vote at the clarity level. Because Grok proved more accurate individually, it contributes five votes (from its five sampled outputs), while Gemini contributes one block vote with weight $w=4$ from its majority clarity label. If the two models already agree on clarity, we keep that label directly. Otherwise, we apply weighted voting over 9 total weighted signals.

Formally, the ensemble decision can be written: Let $m(\cdot)$ be the deterministic evasion-to-clarity mapping, and let $g_i(\ell)$ be Grok’s vote count for evasion label $\ell$ on sample $i$. Let $c_i^{M}$ be Gemini’s clarity label (from Gemini majority evasion), and $w=4$ the Gemini block weight. Stage-1 weighted clarity votes are:
\[
  V_i(c)=\sum_{\ell:\,m(\ell)=c} g_i(\ell) + w\cdot \mathbf{1}[c=c_i^{M}].
\]
If Grok and Gemini already agree on clarity, we keep that label directly. Otherwise, we predict
\[
  \hat{c}_i^{(1)}=\arg\max_{c}V_i(c).
\]

\subsection{Deliberative Complexity Gating (DCG)}

While Stage 1 provides a robust prediction baseline, it struggles significantly with hard, borderline samples. Specifically, on the model-disagreement subset, Stage-1 clarity accuracy drops to 60.0\% (development, 30/50) and 46.9\% (evaluation, 15/32). To address this, we introduce Deliberative Complexity Gating (DCG), a post-hoc and fully unsupervised correction layer applied after Stage 1 voting logic is formed.

\textbf{Cross-Model Signal Discovery.}
To construct an effective gate, we searched for behavioral signals correlated with ambiguity (Ambivalent vs.\ Clear Reply and Clear Non-Reply). As shown in Table~\ref{tab:signal_comparison} (Appendix~\ref{app:main_supp}), we evaluated response-length and self-consistency signals for both models.

Gemini response length is the strongest discriminator among tested signals. Qualitatively, Gemini responses become longer when resolving nuanced evasions, making response length a practical proxy for deliberative difficulty. However, the length condition alone fires on roughly $75\%$ of
the batch. Pairing it with Grok self-consistency adds a precision
filter that restricts overrides to non-unanimous Grok samples. Consequently, correction is applied only when both models indicate uncertainty on the same sample.
The class-wise separation in Gemini response length is visualized in Figure~\ref{fig:box_plot} (Appendix~\ref{app:main_supp}).

\textbf{Gate Logic and Threshold Adaptation.}
Let $L_i$ be the average Gemini response length for sample $i$ ($k{=}5$), and $s_i$ be Grok self-consistency ($k{=}5$). Let $\theta_1$ be the first quartile ($Q_{25}$) of Gemini response lengths computed on the current inference batch:
\[
  \theta_1 = Q_{25}\left(\{L_j\}_{j=1}^{N}\right).
\]
DCG fires when
\[
  o_i = \mathbf{1}\!\left[L_i > \theta_1 \ \land\ s_i < 1.0\right].
\]

If $o_i=1$, DCG overrides only Gemini’s clarity contribution:
\[
  \tilde{c}_i^{M} =
  \begin{cases}
    \text{Ambivalent}, & o_i=1 \\
    c_i^{M},           & o_i=0
  \end{cases}
\]
where $c_i^{M}$ is Gemini’s Stage-1 clarity label. The final Stage-2 prediction is then re-computed with the same weighted voting rule as Stage 1:
\[
  \hat{c}_i^{(2)}=\arg\max_{c}\Big(\sum_{\ell:\,m(\ell)=c} g_i(\ell) + w\cdot\mathbf{1}[c=\tilde{c}_i^{M}]\Big),
\]
with $w=4$, evasion-to-clarity map $m(\cdot)$, and Grok vote counts $g_i(\ell)$.

This keeps DCG transductive and label-free at inference time. The threshold value adapts to each batch distribution, while the percentile choice remains constant across batches.
The threshold $\theta_1 = Q_{25}$ is the first quartile of the batch length distribution and is used as a canonical statistical partition instead of an arbitrary percentile. It reserves the bottom 25\% of response lengths
as a no-fire zone, protecting short responses from
over-firing.
We observed consistent gains on the Development split,
and Appendix~\ref{app:main_supp} (Figure~\ref{fig:percentile_sweep})
reports DCG's full percentile sensitivity.

Figure~\ref{fig:architecture} presents our full two-stage pipeline.

\section{Experimental Setup}

\subsection{Implementation and Dependencies}
We use the official SemEval-2026 Task 6 resources\footnote{\url{https://huggingface.co/datasets/ailsntua/QEvasion}} and the blind evaluation file. Development uses the labeled split ($n=308$), while final evaluation uses the blind set ($n=237$, \texttt{clarity\_task\_evaluation\_dataset.csv}). For evaluation-set reporting, gold labels are read from \texttt{task1\_eval\_labels.txt} and \texttt{task2\_eval\_labels.txt}.

The pipeline is inference-only (no fine-tuning), implemented in Python and uses xAI Grok\footnote{\url{https://x.ai/api}} and Google Gemini\footnote{\url{https://ai.google.dev/gemini-api/docs}} via APIs.
More specifically we use \texttt{gemini-3-flash-preview} and \texttt{grok-4-1-fast-reasoning}. This combination benefits from model diversity, as differences in training and alignment may reduce correlated errors and provide stronger agreement signals for gating and ensemble decisions than using multiple agents of the same model.

\subsection{Inference Hyperparameters}

Both models use self-consistency with $k=5$. Grok uses $\tau=\{0.3,0.5,0.5,0.5,0.5\}$ (first sample as near-deterministic prediction, others for diversity); Gemini uses $\tau=1.0$ (recommended default) with \texttt{thinking\_level=high} and voting weight $w=4$. Stage-2 DCG is post-hoc and API-free. It reads Stage-1 detailed JSON and recomputes final predictions. It uses a batch-adaptive threshold $\theta_1=Q_{25}$ of Gemini response length and a fixed Grok consistency gate $s_i<1.0$.

\subsection{Prompting Strategy}
We use a structured Chain-of-Thought (CoT) \cite{wei2022chain} prompt for evasion classification. The prompt combines (i) taxonomy grounding with explicit definitions for all 9 evasion labels, (ii) contrastive disambiguation rules for confusable classes (especially Dodging vs.\ Implicit vs.\ General), (iii) stepwise decision decomposition (topic match, directness, inferability, refusal, blame-shift, multipart coverage), and (iv) a constrained output schema with mandatory \texttt{FINAL\_LABEL} field. This design improves output parseability and label consistency across self-consistency samples ($k=5$). The full prompt text is provided in Appendix~\ref{app:prompt}.

\subsection{Evaluation Measures}
We report Macro-F1 (primary task metric), accuracy, and per-class precision/recall/F1, computed with scikit-learn\footnote{\url{https://scikit-learn.org/stable/}}.

\section{Results}

\subsection{Evaluation}

In the official SemEval-2026 Task 6 leaderboard, our submission ranked 3rd out of 41 submissions with a Macro-F1 of 0.85, tied also with the second-highest reported result from AsymVerify.
Only TeleAI (Macro-F1\,=\,0.89) reports a higher score as shown in Table~\ref{tab:leaderboard} (Appendix~\ref{app:leaderboard}).

\begin{table}[t]
  \centering
  \small
  \begin{tabular}{lcc}
    \toprule
    \textbf{System Configuration}     & \textbf{Dev F1} & \textbf{Eval F1} \\
    \midrule
    Grok-only ($k=5$)                 & 0.7836          & 0.8264           \\
    Gemini-only ($k=5$)               & 0.7761          & 0.8083           \\
    \midrule
    Ensemble (Stage 1)                & 0.7941          & 0.8122           \\
    \textbf{Ensemble + DCG (Stage 2)} & \textbf{0.8100} & \textbf{0.8505}  \\
    \bottomrule
  \end{tabular}
  \caption{\textbf{Main Results:} Macro-F1 on Development (308 samples) and blind Evaluation (237 samples).}
  \label{tab:main_results}
\end{table}

Table~\ref{tab:main_results} summarizes single-model baselines, Stage-1 ensemble, and Stage-2 DCG.
Stage 1 improves over both single models on Development but underperforms Grok-only on Evaluation, indicating disagreement cases as the main bottleneck.
Adding DCG consistently improves both splits ($+1.60 / +3.83$
macro-F1 points on Development / Evaluation), with parallel accuracy gains (0.8052$\rightarrow$0.8279 and 0.8101$\rightarrow$0.8439). The larger gain on blind evaluation suggests that the gating mechanism transfers well beyond the development distribution (Appendix~\ref{app:a3}).

\subsection{Ablation Study}

To isolate the contribution of each design choice in Stage 1, we run targeted ablations on the Development set, shown in Table~\ref{tab:ablation}. All k-sweep and weight-sweep ablations are computed on Stage-1 (pre-DCG) predictions; DCG is evaluated separately as Stage-2.

\begin{table}[t]
  \centering
  \small
  \begin{tabular}{lc}
    \toprule
    \textbf{Ablation Configuration}  & \textbf{Dev F1}                 \\
    \midrule
    \multicolumn{2}{l}{\textit{Self-Consistency Sweep (Stage 1)}}      \\
    $k=1$ (single pass)              & 0.7845                          \\
    $k=3$                            & \textbf{0.7979}                 \\
    $k=5$ (production)               & 0.7941                          \\
    \midrule
    \multicolumn{2}{l}{\textit{Gemini Weight Sweep ($k=5$)}}           \\
    $w=0$                            & 0.7859                          \\
    $w=1$                            & \textbf{0.8054}                 \\
    $w=2$                            & 0.7909                          \\
    $w=4$ (production)               & 0.7941                          \\
    $w=6$                            & 0.7761                          \\
    \midrule
    \multicolumn{2}{l}{\textit{Multi-Agent Debate vs.\ Static Fusion}} \\
    Stage 1 (static weighted voting) & \textbf{0.7941}                 \\
    Stage 1 + Debate            & 0.7901                          \\
    Multi-agent debate (7-agent, Round 0)     & 0.7773                          \\
    Multi-agent debate (7-agent, Final)       & 0.7656                          \\
    \bottomrule
  \end{tabular}
  \caption{\textbf{Ablations on Development set ( $n=308$).}}
  \label{tab:ablation}
\end{table}

\textbf{Self-Consistency and weights.}
As shown in Table~\ref{tab:ablation}, self-consistency improves over single-pass inference (best standalone Stage-1 at $k=3$, 0.7979), while performance is not strictly monotonic at larger $k$.
For voting weights, $w=1$ gives the best standalone Stage-1 Development score (0.8054), but we keep $w=4$ in the submitted two-stage pipeline to maintain near-parity with Grok (5 vs.\ 4) and to retain disagreement structure for Stage-2 correction.

\textbf{Debate vs.\ static fusion.} We tested whether LLMs could resolve their disagreements through peer-review debate rather than static voting. We applied a debate protocol, asking Grok and Gemini to review each other's rationales on the 50 disagreement-triggered samples. Debate changes 16 predictions, with 8 fixes and 8 degradations (net zero), and reduces macro-F1 from 0.7941 to 0.7901. Behavioral diagnostics show asymmetric adaptation: only-Grok flips = 10, only-Gemini flips = 35. Preservation of already-correct model decisions is also asymmetric (Grok: 22/28, 78.6\%; Gemini: 8/21, 38.1\%).

\textbf{Multi-agent debate comparison.} We recreated the 7-agent homogeneous debate architecture proposed by \citet{hu2025multiagent} to bypass majority voting entirely using \texttt{grok-4-1-fast-reasoning}. A 7-agent debate setup underperforms the static heterogeneous ensemble both before and after debate (0.7773 Round 0, 0.7656 Final, vs.\ 0.7941 for Stage 1), supporting the conclusion that, for this task, model diversity with static fusion is more effective than multi-agent conversational consensus.

\subsection{DCG Justification}

\begin{table}[t]
  \centering
  \small
  \begin{tabular}{lcc}
    \toprule
    \textbf{DCG Configuration}                      & \textbf{Dev F1} & \textbf{Eval F1} \\
    \midrule
    No DCG (Stage 1 only)                           & 0.7941          & 0.8122           \\
    \textbf{Dynamic $\theta_1$ (per-set $Q_{25}$)}  & \textbf{0.8100} & \textbf{0.8505}  \\
    Fixed $\theta_1$ = Dev $Q_{25}$ applied to Eval & —               & 0.8416           \\
    Fixed $\theta_1$ = Eval $Q_{25}$ applied to Dev & 0.8089          & —                \\
    \bottomrule
  \end{tabular}
  \caption{\textbf{Transductive $\theta_1$ adaptation.} Using a threshold from the wrong split degrades performance relative to per-set dynamic adaptation (drop of 0.11--0.89 Macro-F1 points).}
  \label{tab:dcg_threshold}
\end{table}

To test the transductive design of DCG, we compare per-split dynamic thresholding against cross-split fixed threshold transfer as shown in Table~\ref{tab:dcg_threshold}.

The two splits produce different operating thresholds ($\theta_1^{\text{Dev}}=931.0$, $\theta_1^{\text{Eval}}=632.6$), confirming split-dependent response-length scales.
Using a threshold from the wrong split degrades Macro-F1 relative to dynamic per-split $Q_{25}$.
Gate firing is precise for ambiguity detection (81/100 = 81.0\% gold \textit{Ambivalent} on Development; 57/70 = 81.4\% on Evaluation), and its net effect is positive (+7 corrected predictions on Development, +8 on Evaluation).
Sensitivity to the percentile choice is shown in Figure~\ref{fig:percentile_sweep} (Appendix~\ref{app:main_supp}).
Macro-F1 gains over Stage 1
persist across a broad range of percentile values on both splits.

\subsection{Error Analysis}

\textbf{Clear Reply $\leftrightarrow$ Ambivalent boundary dominance (post-DCG).}
On Development ($n=308$), Stage-2 makes 53 clarity errors: 46/53 (86.8\%) are Clear Reply$\leftrightarrow$Ambivalent and 7/53 (13.2\%) are Clear Non-Reply$\leftrightarrow$Ambivalent, with no Clear Reply$\leftrightarrow$Clear Non-Reply errors.
Per-class error rates follow the same pattern: Clear Reply is hardest (25/79, 31.6\%) and Clear Non-Reply the easiest (2/23, 8.7\%).

\textbf{Annotator agreement predicts difficulty.}
Error rate rises from 11.2\% (25/224) on unanimous-label samples to 32.9\% (27/82) on majority-agreement samples.
This is consistent with the task annotation study: clarity-level agreement is higher than evasion-level agreement (Fleiss' $\kappa=0.644$ vs.\ 0.48), with near-perfect agreement for Clear Reply vs.\ Clear Non-Reply ($\kappa=0.97$) and lower agreement when Ambivalent is involved ($\kappa=0.65$--$0.71$) \citep{thomas2024isaidthatdataset}.
Given 72.7\% unanimous samples, the achieved 82.8\% system accuracy indicates robustness under substantial annotation disagreement.
Cross-model agreement vs.\ consistency interaction is shown in Figure~\ref{fig:error_heatmap} (Appendix~\ref{app:main_supp}).

\textbf{CoT paradox.}
For both models step-inconsistent rationales (intermediate reasoning steps point
to a different label than the model's final label) are more accurate than step-consistent ones (70.9\% vs.\ 65.1\% for Grok and 73.3\% vs.\ 63.9\% for
Gemini).
This inconsistency reflects deeper deliberation overriding a
surface-level initial impression. 
The step-inconsistency benefit is \emph{complementary} across models at the class
level.
On the Clear Non-Reply class, Grok answers correctly 94.9\% of the time when
its reasoning steps disagreed with its final label, versus 83.9\% when they
agreed throughout. On the Ambivalent class,
Gemini shows the matching pattern (60.2\% vs.\ 50.7\%).
Each model's accuracy gain from step-inconsistent reasoning lands on a different class and justifies pairing them.

\textbf{Asymmetric handling of one-right disagreements.}
In one-right disagreement cases, the final pipeline preserves Grok-correct decisions much more often than Gemini-correct ones: 27/28 (96.4\%) vs.\ 6/21 (28.6\%).
This asymmetry explains part of the residual error profile and motivates keeping Stage-2 focused on ambiguity correction.

\subsection{Open-Source Replication}

To address the concern that the system depends on proprietary
models, we replicate the DCG mechanism using open-source models\footnote{%
OSS models, identifiers, and providers:
\texttt{kimi-k2.6} (Moonshot AI),
\texttt{deepseek-v4-pro} with max- and high-thinking reasoning effort (DeepSeek),
\texttt{gemma-4-31b-it} (Google),
\texttt{qwen3.6-35b-a3b} (Alibaba).
All run with self-consistency $k=5$.%
},
keeping all hyperparameters fixed at the paper-strict settings to test transferability without per-pairing tuning. The DCG gate combines response length
(Gemini-side) and self-consistency unanimity (Grok-side). For
open-source model selection on the Development split, we
summarise each signal through a single per-role indicator:
Cohen's $d$ between Ambivalent and Clear Reply samples (the
dominant error boundary) for length, and the rate of unanimous
$k=5$ outputs for self-consistency. The two indicators are
complementary to the Cohen's $d$ analysis of
Table~\ref{tab:signal_comparison}: Cohen's $d$ establishes that each
signal type carries class-discriminative information, while
unanimity rate characterises a model's decisiveness profile,
relevant to how often the gate condition $s_i < 1.0$ fires. For
the proprietary baseline both criteria are jointly satisfied. In
the open-source pool they can diverge across candidates. The
proprietary reference values appear in Table~\ref{tab:signal_comparison} (Appendix~\ref{app:main_supp}).

Within the open-source pool (Appendix~\ref{app:main_supp}, Table~\ref{tab:diagnostic}), Kimi K2.6
has the strongest length signal on Development ($d_\text{Dev} =
+0.60$), making it the closest open-source analog of Gemini. Gemma~4 has the highest unanimity rate on Development
($0.67$, exceeding Grok's $0.65$), placing it as a high-decisiveness
candidate for the Grok-role. We replicate the pipeline with Gemma~4
in the Grok-role and Kimi K2.6 in the Gemini-role. The pairing produces
positive transfer on both splits: Stage~1 macro-F1 of 0.7761
(Development) and 0.8317 (Evaluation) is improved by Stage~2 DCG to
0.7787 and 0.8364, corresponding to $\Delta$F1 of $+0.0026$ on
Development and $+0.0047$ on Evaluation. To check whether other OSS-only configurations also produce positive results on both splits,
we tested eight alternatives where the Grok-role is filled by the
OSS model with the strongest Cohen's $d$ for self-consistency (Kimi or Qwen~3.6) instead
of the highest unanimity rate. None yields positive transfer on both splits
(Appendix~\ref{app:oss_alt}, Table~\ref{tab:oss_alt}). The Gemma~4
+ Kimi pairing is therefore the single OSS-only configuration
achieving positive transfer on both splits at paper-strict settings. We do not claim unanimity rate
as a general OSS selection criterion: with a single OSS-only
success, the broader evidence for DCG transfer comes from the
hybrid configurations below.

The same Development-set diagnostic motivates the additional
pairings we test. Four open-source models in
Table~\ref{tab:diagnostic} (Kimi, DeepSeek (max-thinking), DeepSeek (high-thinking), Qwen~3.6)
show consistent positive length separation on Development
($d_\text{Dev} \geq 0.4$), making them candidates for the
length-signal role. Each is paired with Grok at paper-strict
settings, and all four yield positive Stage-2 macro-F1 gains over
Stage 1, reported as $\Delta$F1 (Development / Evaluation): Kimi
$+0.0097 / +0.0152$, DeepSeek (max) $+0.0093 / +0.0124$, DeepSeek
(high) $+0.0031 / +0.0017$, Qwen~3.6 $+0.0041 / +0.0077$. Together with the Gemma~4 + Kimi pairing, these configurations
reproduce DCG's positive transfer across four open-source
families, supporting that the effect is signal-driven and not
tied to any specific proprietary backbone.

\section{Conclusion}

We presented a two-stage heterogeneous system with a novel gating scheme (DCG) for political clarity classification in SemEval-2026 Task 6.
Stage 2 DCG consistently improves over Stage 1, increasing Evaluation Macro-F1 from 0.8122 to 0.8505 and accuracy from 81.0\% to 84.4\%.
Our analysis shows that a clear ceiling remains: 31/308 samples (10.1\%) are misclassified by Grok, Gemini, Stage-1, and Stage-2, and 86.8\% of final errors lie on the Clear Reply $\leftrightarrow$ Ambivalent boundary.
Future work could therefore focus on boundary-specific disambiguation \cite{ren2023self}, and disagreement-aware adaptive weighting \cite{hikal2025msa}.

\section*{Limitations}

Our introduced DCG scheme relies on response-length–based behavioral proxies, which may not always accurately reflect ambiguity across domains or models.
Moreover, DCG requires batch inference, as it depends on using $\theta_1$: the first quartile ($Q_{25}$) of response lengths from the current running batch. Finally, while dependence on proprietary LLMs limits
resource-constrained deployment, our open-source replication demonstrates that DCG transfers to
open-weight backbones, though with reduced magnitude at paper-strict settings.

\bibliography{custom}

\clearpage
\appendix
\makeatletter
\@addtoreset{table}{section}
\@addtoreset{figure}{section}
\makeatother

\renewcommand{\thetable}{\thesection\arabic{table}}
\renewcommand{\thefigure}{\thesection\arabic{figure}}
\onecolumn

\section*{Acknowledgements}

The Gemini experiments (gemini-3-flash-preview) were supported by Google’s Gemini education credits (EDU Credit 423003116).

\section{DCG Analysis}
\label{app:main_supp}

\noindent
This appendix provides supplementary material regarding the DCG gating mechanism.
Each item highlights a different aspect of the DCG design and behavior.

\vspace{0.5cm}
\paragraph{Signal ranking for DCG.}
Table~\ref{tab:signal_comparison} compares candidate behavioral signals for ambiguity detection using Cohen's $d$.
The comparison is stratified by two binary contrasts (Ambivalent vs.\ Clear Reply, and Ambivalent vs.\ Clear Non-Reply) and reported on both splits. Gemini response length is among the strongest and most stable separators across splits, while Grok response length is weaker and self-consistency signals act as complementary cues.

\begin{table}[H]
  \centering
  \small
  \resizebox{0.7\columnwidth}{!}{%
    \begin{tabular}{lcc}
      \toprule
      \textbf{Candidate Signal} & \textbf{$d$ (Amb vs. CR)} & \textbf{$d$ (Amb vs. CNR)} \\
      \midrule
      Gemini response length    & +1.03 / +1.20             & +0.93 / +0.64              \\
      Grok response length      & +0.55 / +0.56             & +0.65 / +0.50              \\
      Grok self-consistency     & -0.41 / -0.66             & -0.54 / -0.73              \\
      Gemini self-consistency   & -0.59 / -0.68             & -0.66 / -0.37              \\
      \bottomrule
    \end{tabular}%
  }
  \caption{\textbf{Effect-size comparison for DCG candidate signals.} Values are reported as Development / Evaluation.}
  \label{tab:signal_comparison}
\end{table}

\begin{figure}[H]
  \centering
  \includegraphics[width=\textwidth]{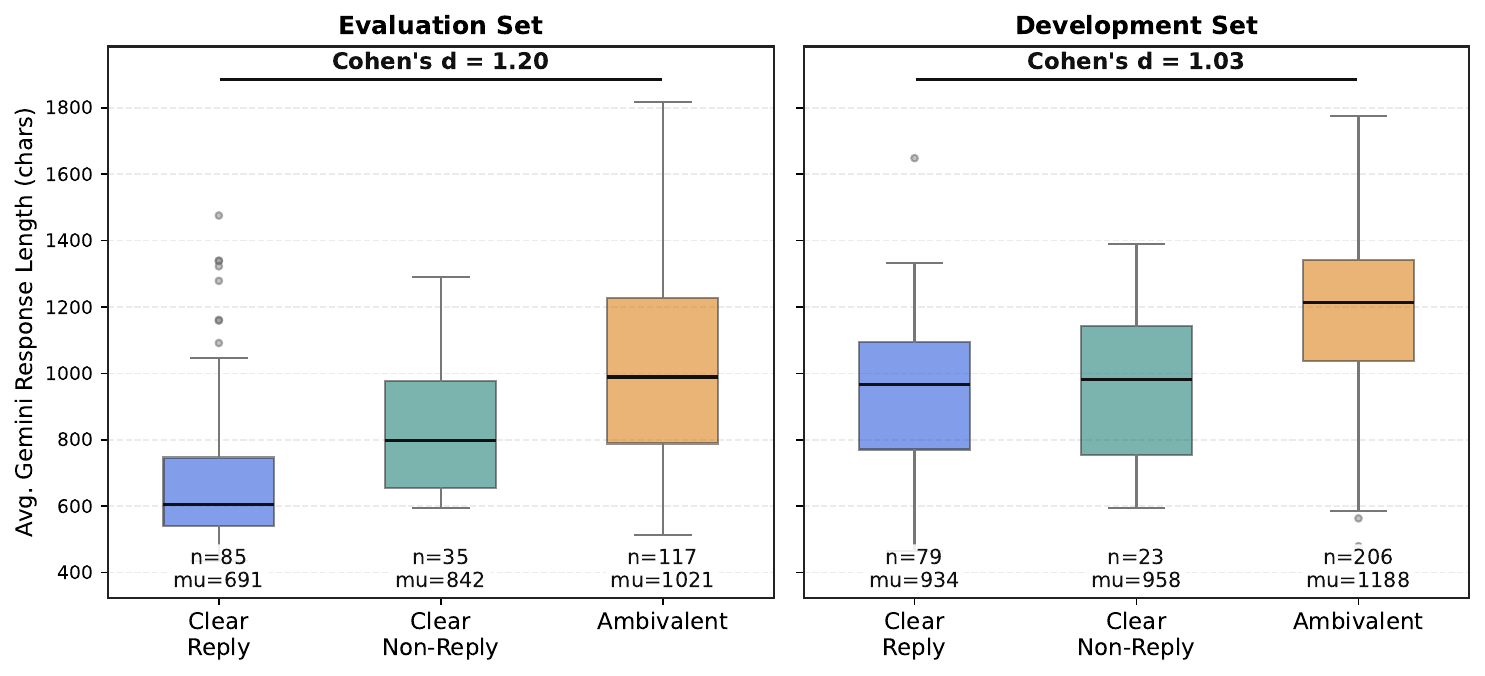}
  \caption{\textbf{Gemini response length by gold clarity class.} Ambivalent samples are generally longer than clear classes on both splits.}
  \label{fig:box_plot}
\end{figure}

\vspace{0.5cm}
\paragraph{Response-length separation by class.}
Figure~\ref{fig:box_plot} provides the class-conditional distributions behind the signal selection.
Ambivalent samples occupy higher response-length ranges than clear classes on both splits, with visible median and spread separation.
This distributional view complements Table~\ref{tab:signal_comparison}: the effect-size advantage of Gemini length is not only numerical but also visually stable across datasets.
\vspace{0.5cm}

\clearpage
\paragraph{Open-source signal profiles.}
Table~\ref{tab:diagnostic} extends the signal analysis of Table~\ref{tab:signal_comparison} to all OSS candidates. Models with high length $d$ (Amb vs.\ CR) match the Gemini role and models with high unanimity rate match the Grok role.

\begin{table}[H]
\centering
\small
\begin{tabular}{lccccc}
\toprule
\textbf{Model} & \textbf{Length} $d$ \textbf{(A vs.\ CR)} & \textbf{Length} $d$ \textbf{(A vs.\ CNR)} & \textbf{Cons.} $d$ \textbf{(A vs.\ CR)} & \textbf{Cons.} $d$ \textbf{(A vs.\ CNR)} & \textbf{Unanim.} \\
\midrule
Grok          & +0.55 / +0.56 & +0.65 / +0.50 & $-0.41$ / $-0.66$ & $-0.54$ / $-0.73$ & 0.65 / 0.69 \\
Gemini        & +1.03 / +1.20 & +0.93 / +0.64 & $-0.59$ / $-0.68$ & $-0.66$ / $-0.37$ & 0.60 / 0.75 \\
\midrule
Kimi K2.6     & +0.60 / +0.82 & +0.22 / +0.23 & $-0.55$ / $-0.65$ & $-0.55$ / $-0.67$ & 0.58 / 0.69 \\
DS-MAX        & +0.45 / +0.87 & +0.62 / +0.39 & $-0.06$ / $-0.52$ & $-0.45$ / $-0.23$ & 0.54 / 0.64 \\
Gemma 4       & +0.27 / +0.50 & +0.66 / +0.33 & $-0.32$ / $-0.60$ & $-0.14$ / $-0.19$ & $0.67$ /  $0.78$ \\
Qwen 3.6      & +0.40 / +0.57 & +0.77 / +0.51 & $-0.49$ / $-0.71$ & $-0.26$ / $-0.29$ & 0.48 / 0.66 \\
DS-HIGH       & +0.41 / +0.72 & +0.45 / +0.21 & $-0.29$ / $-0.79$ & $-0.75$ / $-0.21$ & 0.46 / 0.56 \\
\bottomrule
\end{tabular}
\caption{\textbf{Diagnostic effect-size profiles for open-source DCG candidates.} Cohen's $d$ between gold Ambivalent samples and gold Clear Reply (CR) / Clear Non-Reply (CNR) samples for length and self-consistency, plus the rate of unanimous $k=5$ self-consistency outputs. Values are reported as Development / Evaluation. The two indicators used for OSS selection are Length $d$ (A vs.\ CR) for the Gemini-role and Unanim.\ for the Grok-role; the remaining columns are reported for parallelism with Table~\ref{tab:signal_comparison}. Top block: proprietary baseline. Bottom block: open-source candidates.}
\label{tab:diagnostic}
\end{table}

\paragraph{Alternative OSS-only pairings.}
\label{app:oss_alt}
To check whether other OSS-only configurations also produce positive transfer on both splits, we
evaluated alternative pairings where the Grok-role is occupied by
an open-source model with the strongest self-consistency
separation (Cohen's $d$) rather than the highest unanimity rate. We tested all eight pairings of \{Kimi K2.6, Qwen~3.6\} (the two
OSS models with the strongest self-consistency separation; see
Table~\ref{tab:diagnostic}) in the Grok-role, paired with each of
the remaining four OSS candidates in the Gemini-role.

\begin{table}[H]
\centering
\renewcommand{\arraystretch}{1.0}     
\setlength{\tabcolsep}{12pt}           
\begin{tabular}{llccc}
\toprule
\textbf{Grok-role} & \textbf{Gemini-role} & \textbf{DEV $\Delta$F1} & \textbf{EVAL $\Delta$F1} & \textbf{Verdict} \\
\midrule
\textbf{Gemma 4} & \textbf{Kimi}       & $\mathbf{+0.0026}$ & $\mathbf{+0.0047}$ & $\mathbf{++}$ \\
\midrule
Kimi             & DS-MAX               & $+0.0202$          & $-0.0125$          & $+-$         \\
Kimi             & DS-HIGH              & $+0.0072$          & $-0.0271$          & $+-$         \\
Kimi             & Qwen 3.6             & $-0.0200$          & $-0.0092$          & $--$         \\
Kimi             & Gemma 4              & $-0.0147$          & $-0.0202$          & $--$         \\
Qwen 3.6         & Kimi                 & $-0.0447$          & $-0.0322$          & $--$         \\
Qwen 3.6         & DS-MAX               & $-0.0451$          & $-0.0362$          & $--$         \\
Qwen 3.6         & DS-HIGH              & $-0.0546$          & $-0.0439$          & $--$         \\
Qwen 3.6         & Gemma 4              & $-0.0224$          & $-0.0461$          & $--$         \\
\bottomrule
\end{tabular}%
\caption{\textbf{OSS-only pairings under Cohen's $d$ for self-consistency driven Grok-role selection.} None of the eight Cohen's-$d$-driven alternatives
reproduces DCG's positive transfer on both splits. The first row
(Gemma~4 + Kimi, unanim-driven selection) is repeated for reference.}
\label{tab:oss_alt}
\end{table}

\begin{figure}[H]
  \centering
  \includegraphics[width=0.9\textwidth]{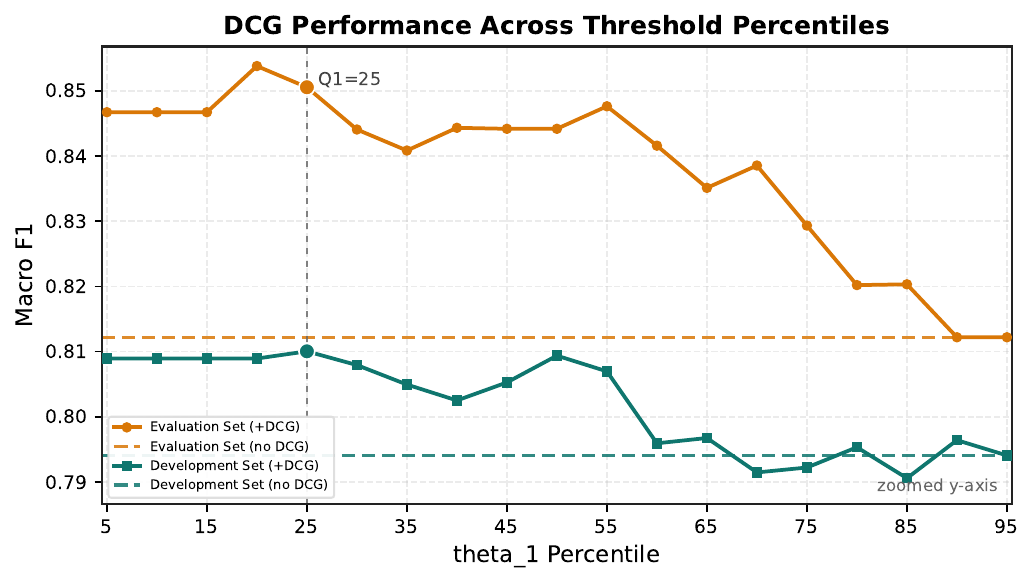}
  \caption{\textbf{DCG sensitivity to percentile choice.} Macro-F1 as $\theta_1$ moves from the 5th to the 95th percentile on Development and Evaluation.}
  \label{fig:percentile_sweep}
\end{figure}

\paragraph{Percentile sensitivity of $\theta_1$.}
Figure~\ref{fig:percentile_sweep} evaluates how Macro-F1 changes as the percentile used for $\theta_1$ varies from 5\% to 95\%.
The main takeaway is robustness: improvements are not confined to a single tuned point, but persist across a broad low-to-mid percentile region.
The selected operating point ($Q_{25}$) remains competitive across splits, while cross-split transfer in the main text shows that per-batch adaptation is still necessary for best performance.

\begin{figure}[H]
  \centering
  \includegraphics[width=\textwidth]{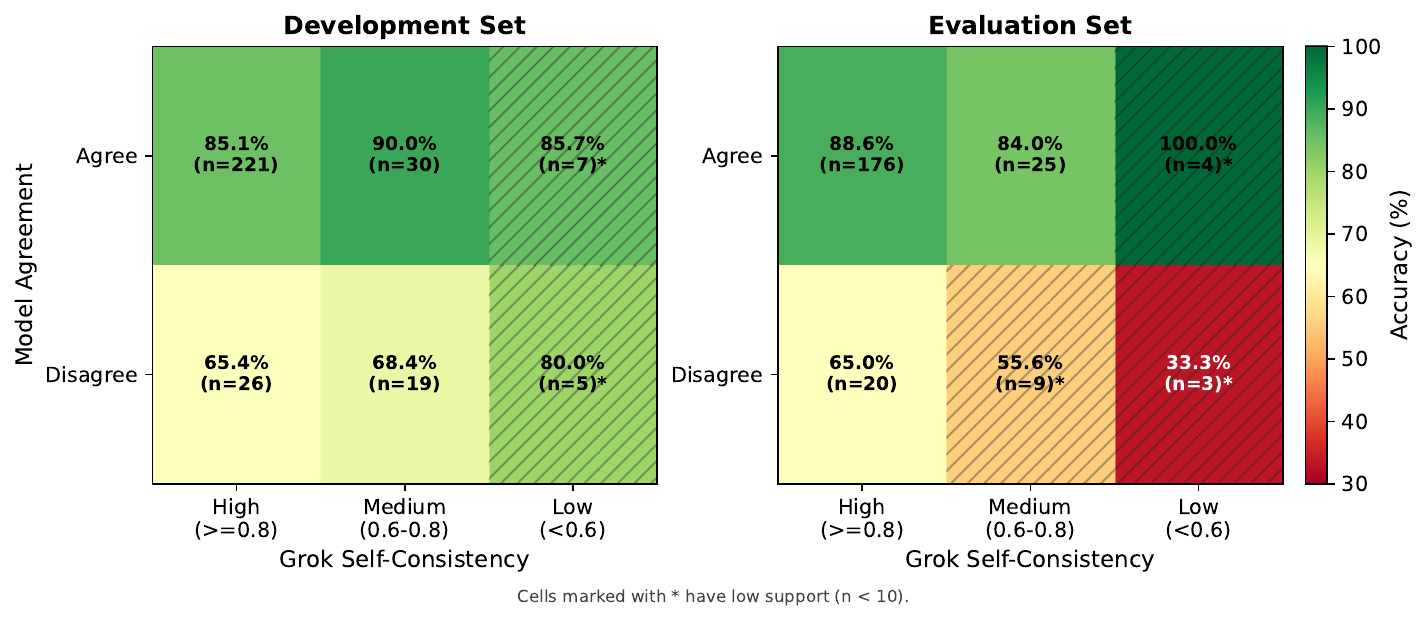}
  \caption{\textbf{Agreement $\times$ consistency heatmap (post-DCG).} Accuracy by Grok--Gemini agreement and Grok self-consistency bins on Development and Evaluation sets.
}
  \label{fig:error_heatmap}
\end{figure}

\paragraph{Error-region structure after DCG.}
Figure~\ref{fig:error_heatmap} decomposes post-DCG accuracy by two axes: inter-model agreement and Grok self-consistency bins.
Cells with agreement are systematically stronger, while disagreement regions (especially medium/low-consistency cells) remain the weakest regions.
This confirms that DCG mitigates, but does not fully eliminate, ambiguity-driven failures at the Clear Reply/Ambivalent boundary.

\vspace{0.5cm}

\twocolumn

\section{Shared-Task Leaderboard}
\label{app:leaderboard}

Table~\ref{tab:leaderboard} shows the top-10 systems on the SemEval-2026 Task 6 Subtask 1 leaderboard.

\begin{table}[H]
\centering
\small
\begin{tabular}{rlc}
\toprule
\textbf{Rank} & \textbf{Team} & \textbf{Macro-F1} \\
\midrule
1            & TeleAI            & 0.89 \\
2            & AsymVerify        & 0.85 \\
\textbf{3}   & \textbf{CSE-UOI}  & \textbf{0.85} \\
4            & Rasende Rakete    & 0.83 \\
5            & Evaluators        & 0.83 \\
6            & YNU-HPCC          & 0.83 \\
7            & moswisarut        & 0.82 \\
8            & tahamunawar       & 0.81 \\
9            & CLaC @ CLARITY    & 0.80 \\
10           & SpinDetector      & 0.80 \\
\bottomrule
\end{tabular}
\caption{Top-10 systems on Subtask 1. Total submissions: 41; median macro-F1: 0.72. Full leaderboard at \url{https://konstantinosftw.github.io/CLARITY-SemEval-2026/}.}
\label{tab:leaderboard}
\end{table}

\section{Full Per-Class Precision / Recall / F1}
\label{app:a3}
Per-class P/R/F1 for Stage-1 and Stage-2
(Tables~\ref{tab:a3_dev},~\ref{tab:a3_eval}): DCG improves
Ambivalent and Clear Non-Reply on Development with a small Clear
Reply trade-off; all three class F1 scores improve on Evaluation.

\begin{table}[H]
\centering\small
\begin{tabular}{lrrr}
\toprule
\textbf{Stage / Class} & \textbf{P} & \textbf{R} & \textbf{F1} \\
\midrule
S1 Ambivalent       & 0.920 & 0.777 & 0.842 \\
S1 Clear Non-Reply  & 0.710 & 0.957 & 0.815 \\
S1 Clear Reply      & 0.641 & 0.835 & 0.725 \\
\midrule
S2 Ambivalent       & 0.870 & 0.874 & 0.872 \\
S2 Clear Non-Reply  & 0.808 & 0.913 & 0.857 \\
S2 Clear Reply      & 0.720 & 0.684 & 0.701 \\
\bottomrule
\end{tabular}
\caption{Per-class clarity metrics, Development ($n{=}308$).}
\label{tab:a3_dev}
\end{table}

\begin{table}[H]
\centering\small
\begin{tabular}{lrrr}
\toprule
\textbf{Stage / Class} & \textbf{P} & \textbf{R} & \textbf{F1} \\
\midrule
S1 Ambivalent       & 0.854 & 0.752 & 0.800 \\
S1 Clear Non-Reply  & 0.806 & 0.829 & 0.817 \\
S1 Clear Reply      & 0.765 & 0.882 & 0.820 \\
\midrule
S2 Ambivalent       & 0.823 & 0.872 & 0.847 \\
S2 Clear Non-Reply  & 0.936 & 0.829 & 0.879 \\
S2 Clear Reply      & 0.842 & 0.812 & 0.826 \\
\bottomrule
\end{tabular}
\caption{Per-class clarity metrics, Evaluation ($n{=}237$).}
\label{tab:a3_eval}
\end{table}

\section{Per-Evasion-Class Model Accuracy}
\label{app:c3}

Tables~\ref{tab:c3_dev} and~\ref{tab:c3_eval} report Grok and Gemini accuracy on each of the 9
evasion labels; $n$ counts can over-
lap across rows due to multi-label gold annota-
tions. 

\begin{table}[H]
\centering\small
\begin{tabular}{lrrr}
\toprule
\textbf{Label} & \textbf{n} & \textbf{Grok} & \textbf{Gemini} \\
\midrule
Explicit      & 115 & 0.774 & 0.861 \\
Implicit      &  99 & 0.748 & 0.667 \\
General       & 113 & 0.637 & 0.611 \\
Dodging       &  96 & 0.531 & 0.531 \\
Deflection    &  51 & 0.647 & 0.490 \\
Partial       &  14 & 0.714 & 0.643 \\
Declining     &  17 & 0.824 & 0.882 \\
Ignorance     &  15 & 0.867 & 0.933 \\
Clarification &   4 & 1.000 & 1.000 \\
\bottomrule
\end{tabular}
\caption{Evasion-level accuracy per model, Development.}
\label{tab:c3_dev}
\end{table}

\begin{table}[H]
\centering\small
\begin{tabular}{lrrr}
\toprule
\textbf{Label} & \textbf{n} & \textbf{Grok} & \textbf{Gemini} \\
\midrule
Explicit      &  84 & 0.881 & 0.976 \\
Implicit      &  23 & 0.522 & 0.565 \\
General       &  19 & 0.632 & 0.579 \\
Dodging       &  94 & 0.362 & 0.575 \\
Deflection    &   0 & --    & --    \\
Partial       &  23 & 0.087 & 0.261 \\
Declining     &  20 & 0.650 & 0.700 \\
Ignorance     &  10 & 1.000 & 0.800 \\
Clarification &   9 & 1.000 & 1.000 \\
\bottomrule
\end{tabular}
\caption{Evasion-level accuracy per model, Evaluation.}
\label{tab:c3_eval}
\end{table}

\section{Stage-2 Clarity Confusion Matrices}
\label{app:confusion}

Tables~\ref{tab:cm_dev} and~\ref{tab:cm_eval} present the post-DCG
(Stage-2) confusion matrices for clarity prediction on Development
and Evaluation. Off-diagonal mass concentrates at the Clear Reply
$\leftrightarrow$ Ambivalent boundary (86.8\% of Development
errors and 78.4\% of Evaluation errors), with no Clear Reply
$\leftrightarrow$ Clear Non-Reply confusions on either split.

\begin{table}[H]
\centering\small
\begin{tabular}{lrrrr}
\toprule
\textbf{Gold $\backslash$ Pred} & \textbf{CR} & \textbf{Amb} & \textbf{CNR} & \textbf{Total} \\
\midrule
Clear Reply       & \textbf{54}  &  25 &   0 &  79 \\
Ambivalent        &  21 & \textbf{180} &   5 & 206 \\
Clear Non-Reply   &   0 &   2 & \textbf{21} &  23 \\
\midrule
Total             &  75 & 207 &  26 & 308 \\
\bottomrule
\end{tabular}
\caption{Stage-2 confusion matrix, Development ($n{=}308$, accuracy $=0.828$). Bold diagonal entries are correct predictions.}
\label{tab:cm_dev}
\end{table}

\begin{table}[H]
\centering\small
\begin{tabular}{lrrrr}
\toprule
\textbf{Gold $\backslash$ Pred} & \textbf{CR} & \textbf{Amb} & \textbf{CNR} & \textbf{Total} \\
\midrule
Clear Reply       & \textbf{69}  &  16 &   0 &  85 \\
Ambivalent        &  13 & \textbf{102} &   2 & 117 \\
Clear Non-Reply   &   0 &   6 & \textbf{29} &  35 \\
\midrule
Total             &  82 & 124 &  31 & 237 \\
\bottomrule
\end{tabular}
\caption{Stage-2 confusion matrix, Evaluation ($n{=}237$, accuracy $=0.844$). Bold diagonal entries are correct predictions.}
\label{tab:cm_eval}
\end{table}

\clearpage
\onecolumn
\section{Prompt Template}
\label{app:prompt}

\paragraph{Design rationale.}
The template is designed to (a) anchor the model in a fixed 9-label taxonomy, (b) reduce class confusion with explicit contrastive rules, and (c) enforce machine-parseable outputs for downstream voting.

\begin{scriptsize}
  \begin{verbatim}
You are an expert at classifying political interview responses for clarity and evasion.
## EVASION TAXONOMY (9 categories):
1. **EXPLICIT**: The answer DIRECTLY states the requested information.
   - Contains yes/no, specific facts, numbers, or clear commitments
   - The EXACT information requested is provided
   - Example: Q: "Will you raise taxes?" A: "No, I will not raise taxes."

2. **IMPLICIT**: The answer is ON-TOPIC and you can INFER the answer from context.
   - MUST be about the SAME TOPIC as the question
   - Not stated directly, but a reasonable listener can deduce the answer
   - Example: Q: "Do you support the bill?" A: "I've always stood with working families on this issue." (implies support)

3. **PARTIAL/HALF-ANSWER**: Multi-part question where SOME parts are answered, others ignored.
   - Only applies to questions with multiple distinct parts
   - Example: Q: "Will you raise taxes and cut spending?" A: "We won't cut spending." (ignores taxes part)

4. **GENERAL**: ON-TOPIC but TOO VAGUE - empty platitudes with NO specifics.
   - MUST be about the SAME TOPIC as the question
   - You CANNOT infer any concrete answer
   - Example: Q: "What will you do about inflation?" A: "We're committed to economic stability."

5. **DODGING**: Answer is OFF-TOPIC - talks about DIFFERENT subject than what was asked.
   - THIS IS THE MOST IMPORTANT CHECK: Does the answer address WHAT WAS ASKED?
   - If the question asks about X but the answer talks about Y -> DODGING
   - Example: Q: "Did you meet the lobbyist?" A: "Let me tell you about our infrastructure plan."
   - Example: Q: "Do you owe an apology?" A: "Let me talk about the PATRIOT Act..." -> DODGING
   - Example: Q: "Is this negotiable?" A: "We must prevent a vacuum..." -> DODGING (different topic)

6. **DEFLECTION**: Stays ON-TOPIC but SHIFTS BLAME to others.
   - Points fingers at opposition, predecessor, external factors
   - MUST still be about the same topic as the question
   - Example: Q: "Why did YOUR project fail?" A: "The previous administration left us a mess."

7. **DECLINING TO ANSWER**: EXPLICIT refusal to answer.
   - "No comment", "I won't discuss", "I can't speak to that"
   - Example: Q: "Can you confirm...?" A: "I'm not going to comment on ongoing investigations."

8. **CLAIMS IGNORANCE**: Says they DON'T KNOW the answer.
   - "I don't know", "I'm not aware", "I'll have to check"
   - Example: Q: "When did this happen?" A: "I don't have that date. I'll get back to you."

9. **CLARIFICATION**: Asks question BACK instead of answering.
   - "What do you mean by...?", "Are you asking about...?"
   - Example: Q: "Was it your decision?" A: "You mean the public fund?"

## CRITICAL DISTINCTION: DODGING vs IMPLICIT vs GENERAL
The #1 mistake is confusing these three. Here's the key test:

| Question               | If answer talks about...                     | Label                    |
|------------------------|----------------------------------------------|--------------------------|
| "Will you do X?"       | Something OTHER than X                       | **DODGING** (off-topic)  |
| "Will you do X?"       | X, but in a way you can infer the answer     | **IMPLICIT**             |
| "Will you do X?"       | X, but too vague to infer anything           | **GENERAL**              |

**DODGING examples (answer talks about DIFFERENT subject):**
- Q: "Do you owe an apology?" A: "Let me discuss the PATRIOT Act..." -> DODGING (apology != PATRIOT Act)
- Q: "Is this negotiable?" A: "We can't create a vacuum for Hezbollah..." -> DODGING (negotiable != vacuum)
- Q: "Do walls feel like closing in?" A: "Senator Warner said..." -> DODGING (feelings != Senator Warner)

**IMPLICIT examples (on-topic, inferable):**
- Q: "Do you support the bill?" A: "I've always supported working families." -> IMPLICIT (on-topic, implies yes)

**GENERAL examples (on-topic, no inference possible):**
- Q: "What's your plan?" A: "We will use all tools available." -> GENERAL (on-topic, but too vague)

## YOUR TASK:
Classify the given Q&A pair. FIRST check if the answer is ON-TOPIC or OFF-TOPIC.

## OUTPUT FORMAT (follow exactly):
STEP1_QUESTION_TOPIC: <What SPECIFIC topic/subject does the question ask about?>
STEP2_ANSWER_TOPIC: <What topic/subject does the answer ACTUALLY discuss?>
STEP3_TOPIC_MATCH: <Do these topics MATCH? YES = on-topic, NO = off-topic (likely DODGING)>
STEP4_DIRECT_CHECK: <If topics match: Is there a DIRECT yes/no or specific answer? If yes -> Explicit>
STEP5_INFERENCE_CHECK: <If topics match: Can you INFER an answer? If yes -> Implicit. If no -> General>
STEP6_REFUSAL_CHECK: <Is there explicit refusal, ignorance claim, or clarification request? -> Clear Non-Reply>
STEP7_BLAME_CHECK: <If topics match: Does the answer shift blame to others? If yes -> Deflection>
STEP8_MULTI_PART_CHECK: <If multi-part question, are ALL parts addressed? If some missing -> Partial>
FINAL_LABEL: <One of: Explicit, Implicit, Partial/half-answer, General, Dodging, Deflection,
              Declining to answer, Claims ignorance, Clarification>
CONFIDENCE: <1-5, where 5 is highest confidence>
\end{verbatim}
\end{scriptsize}

\clearpage
\twocolumn

\end{document}